\documentclass[10pt,twocolumn,letterpaper]{article}

\usepackage{cvpr}
\usepackage{times}
\usepackage{epsfig}
\usepackage{graphicx}
\usepackage{amsmath}
\usepackage{amssymb}
\usepackage{xcolor}
\usepackage{subfig}
\usepackage{mwe}
\usepackage{booktabs}
\usepackage{multirow}
\usepackage{adjustbox}

\usepackage[breaklinks=true,bookmarks=false]{hyperref}

\cvprfinalcopy 

\def\cvprPaperID{16} 

\begin{document}

\title{Explaining  Autonomous Driving by Learning End-to-End Visual Attention}

\author{Luca Cultrera ~~~ Lorenzo Seidenari ~~~ Federico Becattini ~~~ Pietro Pala ~~~ Alberto Del Bimbo\\
MICC, University of Florence\\
{\tt\small name.surname@unifi.it}
}

\maketitle
\thispagestyle{empty}

\begin{abstract}
    Current deep learning based autonomous driving approaches yield impressive results also leading to in-production deployment in certain controlled scenarios. One of the most popular and fascinating approaches relies on learning vehicle controls directly from data perceived by sensors. This end-to-end learning paradigm can be applied both in classical supervised settings and using reinforcement learning. Nonetheless the main drawback of this approach as also in other learning problems is the lack of explainability. Indeed, a deep network will act as a black-box outputting predictions depending on previously seen driving patterns without giving any feedback on why such decisions were taken.
    
    While to obtain optimal performance it is not critical to obtain explainable outputs from a learned agent, especially in such a safety critical field, it is of paramount importance to understand how the network behaves. This is particularly relevant to interpret failures of such systems.
    
    In this work we propose to train an imitation learning based agent equipped with an attention model. The attention model allows us to understand what part of the image has been deemed most important. Interestingly, the use of attention also leads to superior performance in a standard benchmark using the CARLA driving simulator.

\end{abstract}
\section{Introduction}
Teaching an autonomous vehicle to drive poses a hard challenge.
Whereas the problem is inherently difficult, there are several issues that have to be addressed by a fully fledged autonomous system. First of all, the vehicle must be able to observe and understand the surrounding scene. If data acquisition can rely on a vast array of sensors \cite{geiger2013vision}, the understanding process must involve sophisticated machine learning algorithms \cite{chen2017rethinking, he2017mask, berlincioni2019road, mur2017orb, wang2017pedestrian}. Once the vehicle is able to perceive the environment, it must learn to navigate under strict constraints dictated by street regulations and, most importantly, safety for itself an others.
To learn such a skill, autonomous vehicles can be instructed to imitate driving behaviors by observing human experts \cite{Alpher22, Alpher24, Alpher30, Alpher03, Alpher09}. Simulators are often used to develop and test such models, since machine learning algorithms can be trained without posing a hazard for others \cite{Alpher20}.
In addition to learning to transform visual stimuli into driving commands, a vehicle also needs to estimate what other agents are or will be doing \cite{rhinehart2019precog, becattini2019vehicle, ma2019trafficpredict, marchetti2020memnet}.

Even when it is able to drive correctly, under unusual conditions an autonomous vehicle may still commit mistakes and cause accidents. When this happens, it is of primary importance to assess what caused such a behavior and intervene to correct it.
Therefore, explainability in autonomous driving assumes a central role, which is often not addressed as it should. Common approaches in fact, although being effective in driving tasks, do not offer the possibility to inspect what generates decisions and predictions.
In this work we propose a conditional imitation learning approach that learns a driving policy from raw RGB frames and exploits a visual attention module to focus on salient image regions. This allows us to obtain a visual explanation of what is leading to certain predictions, thus making the model interpretable.
Our model is trained end-to-end to estimate steering angles to be given as input to the vehicle in order to reach a given goal. Goals are given as high-level commands such as drive straight or turn at the next intersection. Since each command reflects in different driving patterns, we propose a multi-headed architecture where each branch learns to perform a specialized attention, looking at what is relevant for the goal at hand.
The main contributions of the paper are the following:
\begin{itemize}
\item We propose an architecture with visual attention for driving an autonomous vehicle in a simulated environment. To the best of our knowledge we are the first to explicitly learn attention in an autonomous driving setting, allowing predictions to be visually explained.
\item We show that the usage of visual attention, other than providing an explainable model, also helps to drive better, obtaining state of the art results on the CARLA driving benchmark \cite{Alpher20}.
\end{itemize}

The work is described as follows: In Section \ref{related} the related works are described to frame our method in the current state of the art; in Section \ref{method} the method is shown using a top-down approach, starting from the whole architecture and then examining all the building blocks of our model. In Section \ref{results} results are analyzed and a comparison with the state of the art is provided. Finally, in Section \ref{conclusion} conclusions are drawn.


\section{Related Works}
\label{related}
Our approach can be framed in the imitation learning line of investigation of autonomous driving. Our model improves the existing imitation learning framework with an attention layer. Attention in imitation learning is not extensively used,  therefore in the following we review previous work in imitation learning and end-to-end driving and attention models.
\subsection{Imitation Learning \& end-to-end driving}
One of the key approaches for learning to execute complex tasks in robotics is to observe demonstrations, performing  so called imitation learning~\cite{Alpher05, Alpher06}.
Bojarski et al. \cite{Alpher02} and Codevilla et al. \cite{Alpher03} were the first to propose a successful use of imitation learning for autonomous driving.

Bojarski et al \cite{Alpher02} predict steering commands for lane following and obstacle avoiding tasks. Differently, the solution proposed by Codevilla et al. \cite{Alpher03} performs conditional imitation learning, meaning that the network emits predictions conditioned on high level commands. A similar branched network is proposed by Wang et al. \cite{Alpher28}. Liang et al. \cite{Alpher24} instead use reinforcement learning to perform conditional imitation learning.

Sauer et al. \cite{Alpher23}, instead of directly linking perceptual information to controls, use a low-dimensional intermediate representation, called affordance, to improve generalization.  A similar hybrid approach, proposed by  Chen et al. \cite{Alpher27}, maps input images to a small number of perception indicators relating to affordance environment states.

Several sensors are often available and simulators~\cite{Alpher03} provide direct access to depth and semantic segmentation. Such information is leveraged by several approaches~\cite{Alpher22, Alpher25, Alpher26}.
Recently Xiao et al. \cite{Alpher22} improved over \cite{Alpher03, Alpher25} adding depth information to RGB images obtaining better performances. Li et al. \cite{Alpher25} exploit depth and semantic segmentation at training time performing multi-task learning while in \cite{Alpher22} depth is considered as an input to the model. Xu et al. \cite{Alpher26} show that it is possible to learn from a crowd sourced video dataset using multiple frames with an LSTM. They also demonstrate that predicting semantic segmentation as a side task improves the results. Multi-task learning proves effective also in \cite{Alpher32} for end-to-end steering and throttle prediction. Temporal modelling is also used by \cite{Alpher29} and \cite{Alpher30} also using Long Short-Term Memory Networks.

Zhang and Cho \cite{Alpher09} extend policy learning allowing a system to gracefully fallback into a safe policy avoiding to fall into dangerous states. 
\subsection{Attention models} 
Attention has been used in classification, recognition and localization tasks, as in \cite{Alpher34, Alpher35, Alpher36}. The authors of \cite{Alpher37} propose an attention model for object detection and counting on drones, while \cite{Alpher38} uses attention for small object detection. Other examples of attention models used for image classification are  \cite{Alpher39, Alpher40, Alpher41, Alpher42}.  Often, attention-based networks are used for  video summarization \cite{Alpher11, Alpher12} or Image Captioning task, as in the work of Anderson et al. \cite{Alpher13} or Cornia et al. 
\cite{Alpher14}.
There are some uses of attention based models in autonomous driving \cite{Alpher15, Alpher16}.  Attention  has been used to improve interpretability in autonomous driving by Jinkyu and Canny \cite{Alpher15}. Salient regions extracted from a saliency model are used to condition the network output by weighing network feature maps with such attention weights.  Chen et al. \cite{Alpher16}  adopt a  brain inspired cognitive model. Both are multi-stage approaches in which attention is not computed in an end-to-end learning framework.  

Our approach differs from existing ones since we learn attention directly during training in an end-to-end fashion instead of using an external source of saliency to weigh intermediate network features.
We use proposal regions that our model associates with a probability distribution to highlight salient regions of the image. This probability distribution indicates how well the corresponding regions predict steering controls.


\section{Method}
\label{method}
To address the autonomous driving problem in an urban environment, we adopt an Imitation Learning strategy to learn a driving policy from raw RGB frames.
\subsection{Imitation learning}\label{il}
\begin{figure}[t]
\centering
\includegraphics[width=\columnwidth]{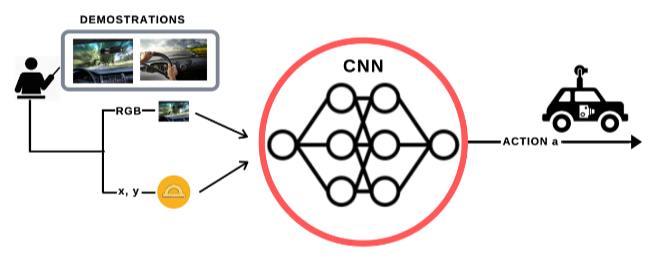}
\caption{Imitation learning}
\label{fig:fig1}
\end{figure}

Imitation learning is the process by which an agent attempts to learn a policy \textit{$\pi$} by observing a set of demonstrations \textit{D}, provided by an expert \textit{E} \cite{Alpher04}. Each demonstration is characterized by an input-output couple $D = {(z_i,a_i)}$, where z$_{i}$ is the i-th state observation and a$_{i}$ the action performed in that instant.
The agent does not have direct access to the state of the environment, but only to its representation.
In the simplest scenario, an imitation learning process, is a direct mapping from state observations to output actions. In this case the policy to be learned is obtained by the mapping \cite{Alpher05}:
\begin{equation}{\pi: Z \rightarrow A}
\end{equation}
where \textit{Z} represents the set of observations and \textit{A} the set of possible actions.
In an autonomous driving context, the expert \textit{E} is a driver and the policy is \textit{``safe driving"}.
Observations are RGB frames describing the scene captured by a vision system and actions are driving controls such as throttle and steering angle. Therefore, the imitation learning process can be addressed through convolutional neural networks (CNN). Demonstrations are a set of \textit{(frame, driving-controls)} pairs, acquired during pre-recorded driving sessions performed by expert drivers. The high level architecture of this approach is shown in Figure \ref{fig:fig1}.

\subsection{Approach}
Our Imitation Learning strategy consists in learning a driving policy that maps observed frame to the steering angle needed by the vehicle to comply with a given high level command.
We equip our model with a visual attention mechanisms to make the model interpretable and to be able to explain the estimated maneuvers.

The model, which is end-to-end trainable, first extracts features maps from RGB images with a fully convolutional neural network backbone. A regular grid of regions, obtained by sliding boxes of different scales and aspect ratios over the image, selects Regions of Interest (RoI) from the image. Using a RoI Pooling strategy, we then extract a descriptor for each region from the feature maps generated by the convolutional block. 
An attention layer assigns weights to each RoI-pooled feature vector. These are then combined together with a weighted sum to build a descriptor that is fed as input to the last block of the model, which includes dense layers of decreasing size and that emits steering angle predictions.

\begin{figure*}[h]
\centering
\includegraphics[width=\textwidth ]{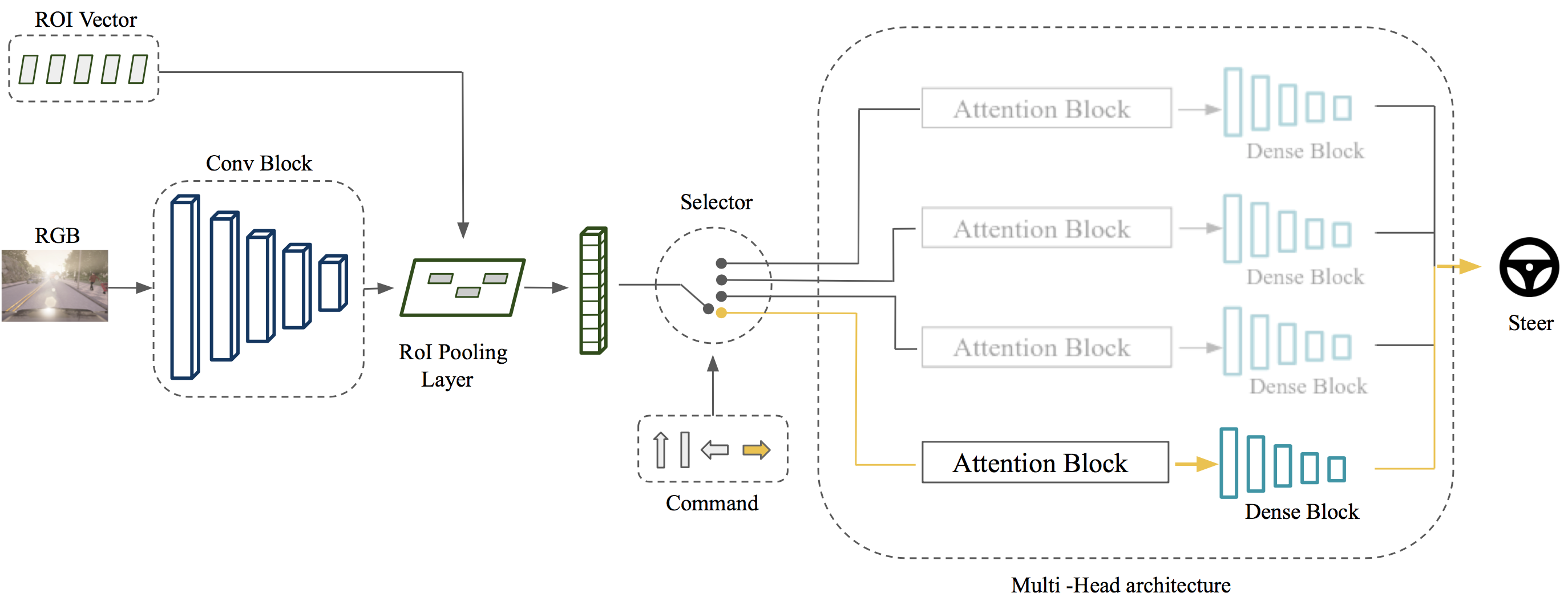}
\caption{Architecture. A convolutional backbone generates a feature map from the input image. A fixed grid of Regions Of Interest is pooled with ROI pooling and weighed by an attention layer. Separate attention layers are trained for each high level command in order to focus on different regions and output the appropriate steering angle.}
\label{fig:fig4}
\end{figure*}

The system is composed of a shared convolutional backbone, followed by a multi-head visual attention block and a final regressor block.
The proposed multi-head architecture is shown in Figure \ref{fig:fig4}.
The different types of high-level commands that can be given as input to the model include:
\begin{itemize}
\item \textit{Follow Lane}: follow the road
\item \textit{Go straight}: choose to go straight to an intersection
\item \textit{Turn Left}: choose to turn left to an intersection
\item \textit{Turn right}: choose to turn right to an intersection
\end{itemize}

In the following, we present in detail each module that is employed in the model.

\subsection{Shared Convolutional Backbone}\label{conv}
The first part of the model takes care of extracting features from input images. Since our goal is to make the model interpretable, in the sense that we want to understand which pixels contribute more to a decision, we want feature maps to preserve spatial information. This is made possible using a fully convolutional structure.
In particular, this sub-architecture is composed of 5 convolutional layers: 
the first three layers have respectively 24, 36 and 48 kernels of size $5 \times 5$ and stride 2 and are followed by two other layers both with 64 $3 \times 3$ filters with stride 1.
All convolutional layers use a Rectified Linear Unit (Relu) as activation function. Details of the convolutional backbone are summarized in Table \ref{table:table1}.
The convolutional block is followed by a RoI pooling layer which maps regions of interest onto the feature maps.

\begin{table}[t]
\centering
\begin{tabular}{ |c|c|c|c|c| } 
\hline
\# & Output dim & Channels & Kernel size & Stride \\
\hline
1 & $298 \times 130 \times 24$ & 24 & 5 & 2  \\ 
2 & $147 \times 63 \times 36$ & 36 & 5 & 2  \\ 
3 & $72 \times 30 \times 48$ & 48 & 5 & 2  \\ 
4 & $70 \times 28 \times 64$ & 64 & 3 & 1  \\ 
5 & $68 \times 26 \times 64$ & 64 & 3 & 1  \\ 
\hline
\end{tabular}
\caption{Convolutional backbone architecture. The size of the input images is $600 \times 264$.}
\label{table:table1}
\end{table}

\subsection{Region Proposals}
\label{RoiPooling}
The shared convolutional backbone, after extracting features from input images, finally conveys into a RoI pooling layer \cite{Alpher17}.
For each Region of Interest, which can exhibit different sizes and aspect ratios, the RoI pooling layer extracts a fixed-size descriptor $r_i$ by dividing the region into a number of cells on which a pooling function is applied. Here we adopt the max-pooling operator over $4 \times 4$ cells.



\begin{figure}[t]
\centering
\includegraphics[width = .99\columnwidth]{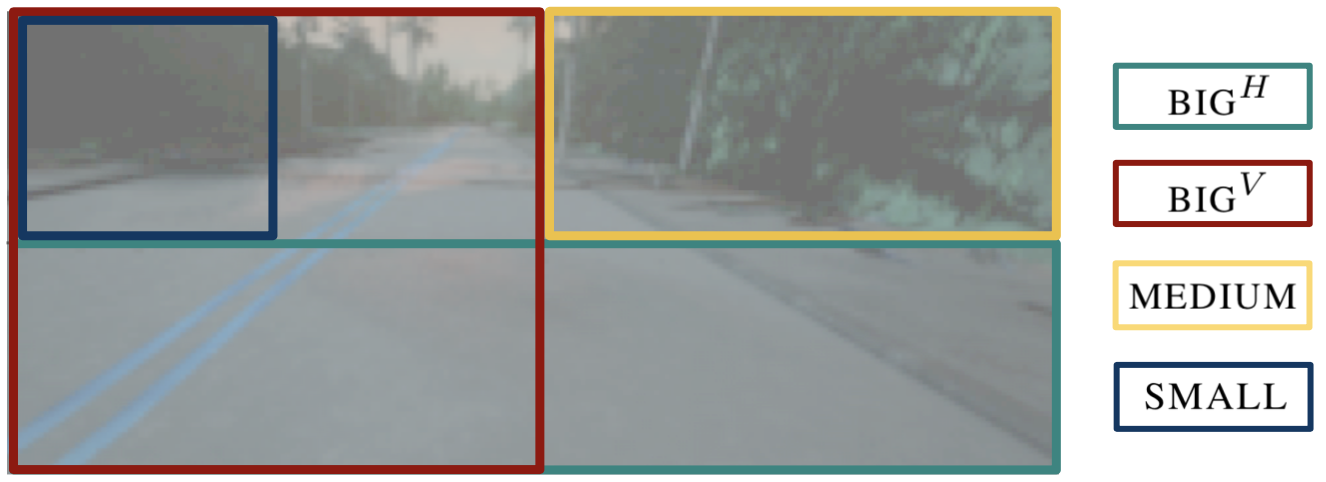}
\caption{Four sliding windows are used to generate a multi-scale grid of RoIs. Colors indicate the box type: $\textsc{big}^V$ (red), $\textsc{big}^H$ (green), $\textsc{medium}$ (yellow) $\textsc{small}$ (blue).}
\label{fig:fig2}
\end{figure}

RoI generation is a fundamental step in our pipeline since extracting good RoIs allows the attention mechanism, explained in Section \ref{attention}, to correctly select salient regions of the image.
To extract RoIs from an image of size $H \times W$ we use a multi-scale regular grid. The grid is obtained by sliding multiple boxes on the input image with a chosen step size. For this purpose we used four sliding windows with different strides as explained in Figure \ref{fig:fig2}:
\begin{itemize}
\item $\textsc{big}^H$: a horizontal box of size $H/2 \times W$ that covers the whole width of the image, ranging from top to bottom with a 60px vertical stride.
\item $\textsc{big}^V$: a vertical box of size $H \times W/2$ with horizontal stride equal to $W/2$, therefore yielding two regions dividing the image into a left and right side.
\item $\textsc{medium}$: a box of size $H/2 \times W/2$ covering a quarter of the image. The sliding window is applied on the upper and lower half of the image with an horizontal stride of 60px.
\item $\textsc{small}$: a square box of size $H/2 \times W/4$, applied with stride 30px in both directions over the image.
\end{itemize}

We identify each window type as $\textsc{big}$, $\textsc{medium}$ and $\textsc{small}$ to address their scale and we use the $\textsc{h}$ and $\textsc{v}$ superscripts to respectively refer to the horizontal and vertical aspect ratios.

The four box types are thought to take into account different aspects of the scene. The first scale ($\textsc{big}^H$ and $\textsc{big}^V$) is coarse and follows the structural elements in the scene (e.g. vertical for traffic signs or buildings and horizontal for forthcoming intersections). The remaining scales instead focus on smaller elements such as vehicles or distant turns.

In total we obtain a grid of 48 regions: 2 $\textsc{big}^V$, 6 $\textsc{big}^H$, 8 $\textsc{medium}$ and 32 $\textsc{small}$.
Note that despite having a fixed grid may look as a limitation, this is necessary to ensure that the model understands the spatial location of each observed region. Furthermore it allows to take into account all regions at the same time to generate the final attention, which is more effective than generating independent attentions for each region. This aspect is further discussed in Section \ref{sec:ablation}.

\subsection{Visual Attention}\label{attention}
Visual attention allows the system to focus on salient regions of the image. In an autonomous driving context, explicitly modeling saliency can help the network to take decisions considering relevant objects or environmental cues, such as other vehicles or intersections.
For this purpose, the attention layer must be able to weigh image regions by estimating a probability distribution over the calculated set of RoI-pooled features.
We learn to estimate visual attention directly from the data, training the model end-to-end to predict the steering angle of the ego-vehicle in a driving simulator. Therefore the attention layer is trained to assign weights to image regions by relying on the relevance of each input in the driving task.
In order to condition attention on high level driving commands, we adopt a different head for each command. Each head is equipped with an attention block structured as follows.

At first, region features $r_i$ obtained by RoI-pooling are flattened and concatenated together in a single vector $r$. This is then fed to a fully connected layer that outputs a logit for each region. Each logit is then normalized with a softmax activation to be converted into a RoI weight $\alpha = \alpha_1, ..., \alpha_R$, where $R$ is the number of regions:

\begin{equation}
    \alpha = Softmax(r \cdot W_a + b_a)
\end{equation}

Here $W_a$ and $b_a$ are the weights and biases of the linear attention layer, respectively. The softmax function helps the model to focus only on a small subset of regions, since it boosts the importance of the regions with the highest logits and dampens the contribution of all others.
The final attention feature $r_a$ is obtained as a weighted sum of the region features $r_i$:

\begin{equation}
    r_a = \sum_{i=1}^{R} r_i \cdot \alpha_i
\end{equation}

\begin{figure}[t]
\centering
\includegraphics[width=\columnwidth]{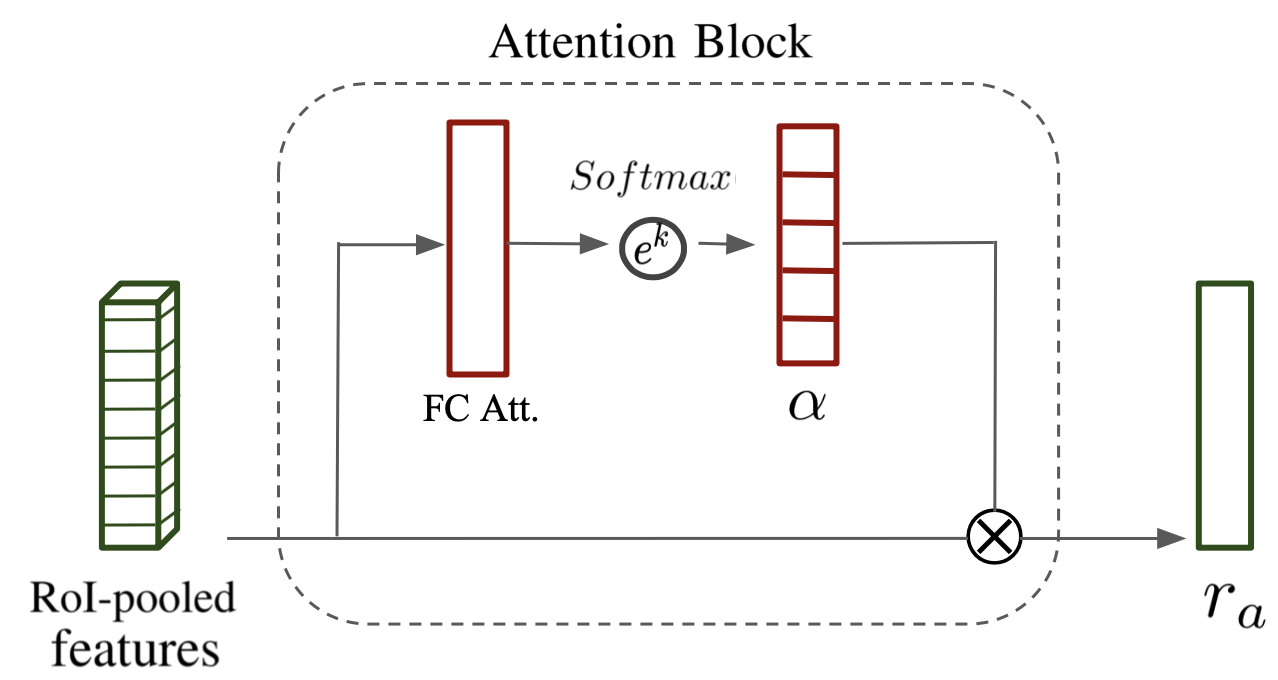}
\caption{Attention Block. A weight vector $\alpha$ is generated by a linear layer with softmax activation. The final descriptor is obtained as a weighted sum of RoI-pooled features.}
\label{fig:attention_block}
\end{figure}

The architecture of the attention block adopted in each head of the model is shown in Figure \ref{fig:attention_block}.


\subsection{Multi-head architecture}
The model is requested to output a steering angle based on the observed environment and a high level command. High level commands encode different behaviors, such as moving straight or taking a turn, which therefore entail different driving patterns. To keep our architecture as flexible and extendable as possible, we use a different prediction head for each command, meaning that additional heads could be easily plugged in to address additional high level commands. Moreover, multi-head architectures have been shown to outperform single-headed models \cite{Alpher03, Alpher23}. Each command has its own specialized attention layer, followed by a dense block to output predictions. This has also the benefit of increasing explainability, since we can generate different attention maps conditioned on different commands.

The weighted attention feature $r_a$, produced by the attention layer, 
is given as input to a block of dense layers of decreasing size, i.e. 512, 128, 50, 10, respectively. Finally, a last dense layer predicts the steering angle required to comply with the given high level command.

The high level command is provided to the model along with the input image and acts as a selector to determine which head to use. Therefore, at training time, the error is backpropagated only through the head specified by the current command and then through the shared feature extraction backbone, up to the input in an end-to-end fashion.


\subsection{Training details}
As input for the model we use 600$\times$264 images captured by a camera positioned in front of the vehicle.
Throughout training and testing we provide a constant throttle to the vehicle, with an average speed of $10Km/h$ as suggested in the CARLA driving benchmark \cite{Alpher20}.
For each image, a steering angle \textit{s} represents the action performed by the driver. 
Therefore the loss function is calculated using Mean Squared Error (MSE) between the predicted value and the ground truth steer \eqref{eq:10}:

\begin{equation}
\label{eq:10}
    \mathcal{L} = \Vert s_{GT} - s_p \Vert^2 
\end{equation}

Where $s_{GT}$ and $s_p$ represent respectively ground truth and predicted steer values.
The model is trained using the Adam optimizer \cite{Alpher19} for 20 epochs with batch size of 64 and initial learning rate of 0,0001.

\subsection{Dataset}
For training and evaluating our model, we use data from the CARLA simulator \cite{Alpher20}.
CARLA is an open source urban driving simulator, that can be used to test and evaluate autonomous driving agents. The simulator provides two realistically reconstructed towns with different day time and weather conditions.

The authors of the simulator, also released a dataset for training and testing \cite{Alpher03}. The dataset uses the first town to collect training images and measurements and the second town for evaluation only.
The training set is composed by data captured in four different weather conditions for about 14 hours of driving. For each image in the dataset, a vector of measurements is provided which includes values for steering, accelerator, brake, high level command and additional information that can be useful for training an autonomous driving system such as collisions with other agents.

To establish the capabilities of autonomous agents, CARLA also provides a driving benchmark, in which agents are supposed to drive from a starting anchor point on the map to an ending point, under different environmental conditions.
The benchmark is goal oriented and is composed of several episodes divided in four main task:
\begin{enumerate}
\item \textit{Straight}: drive in a straight road.
\item \textit{One turn}: drive performing at least one turn.
\item \textit{Navigation}: driving in an urban scenario performing several turns.
\item \textit{Navigation Dynamic}: same scenario as \textit{Navigation}, including pedestrians and other vehicles.
\end{enumerate}
For each task, 25 episodes are proposed in several different weather conditions (seen and unseen during training). Two towns are used in the benchmark: \textit{Town1}, i.e.~the same town used for training the network, and \textit{Town2}, which is used only at test time. In total the benchmark consists of 1200 episodes, 600 for each town.


\newcommand{\depth}{\blacklozenge}
\newcommand{\segm}{\mathsection}
\newcommand{\temporal}{\dagger}

\begin{table*}[t]
\begin{adjustbox}{max width=.8\linewidth,center}
\begin{tabular}{@{}ccccc||cccccc@{}}
\toprule
Method & MP \cite{Alpher20} & MT \cite{Alpher25} & CAL \cite{Alpher23} & EF \cite{Alpher22} & RL \cite{Alpher20} & IL \cite{Alpher20, Alpher03} & EF-RGB \cite{Alpher22} & CIRL \cite{Alpher24} & Ours no Att. & Ours \\ \midrule 
Additional data & S & S+D & T & D & - & - & - & - & - & - \\ \midrule
Success Rate & 69	& 83	& 84	& \textbf{92}	& 27	& 72 & 75	& 82	& 69 &	\textbf{84} \\ \bottomrule
\end{tabular}
\end{adjustbox}
\caption{Success Rate averaged across all tasks. For each method we show whether additional data is used other than a single RGB frame: temporal sequence of frames (T), semantic segmentations (S), depth (D).}
\label{table:average}
\end{table*}

\begin{table*}[t]
	\begin{adjustbox}{max width=.8\linewidth,center}
		\begin{tabular}{@{}ccccc||cccccc@{}}
			\toprule
			\multicolumn{11}{c}{Training conditions}\\ \midrule 
			Task & $MP^{\segm}$ \cite{Alpher20} & $MT^{\segm \depth}$ \cite{Alpher25} & $CAL^{\temporal}$ \cite{Alpher23} & $EF^{\depth}$ \cite{Alpher22} & RL \cite{Alpher20} & IL \cite{Alpher20, Alpher03} & EF-RGB \cite{Alpher22} &  CIRL \cite{Alpher24} & Ours no Att. & Ours \\ \midrule
			Straight & 98 & 98 & \textbf{100} & \underline{99} & 89 & 95 & 96 & \underline{98} & \textbf{100} & \textbf{100}\\
			One turn & 82 & 87 & \underline{97} & \textbf{99} & 34 & 89 & \underline{95} & \textbf{97} & 91 & \underline{95}\\
			Navigation & 80 & \underline{81} & \textbf{92} & \textbf{92} & 14 & 86 & 87 & \textbf{93} & 80 & \underline{91}\\
			Nav. Dynamic & 77 & 81 & \underline{83} & \textbf{89} & 7 & 83 & \underline{84} & 82 & 79 & \textbf{89} \\ \bottomrule
		\end{tabular}
	\end{adjustbox}
	\begin{adjustbox}{max width=.8\linewidth,center}
		\begin{tabular}{@{}ccccc||cccccc@{}}
			\toprule
			\multicolumn{11}{c}{New weather}\\ \midrule
			Task & $MP^{\segm}$ \cite{Alpher20} & $MT^{\segm \depth}$ \cite{Alpher25} & $CAL^{\temporal}$ \cite{Alpher23} & $EF^{\depth}$ \cite{Alpher22} & RL \cite{Alpher20} & IL \cite{Alpher20, Alpher03} & EF-RGB \cite{Alpher22} & CIRL \cite{Alpher24} & Ours no Att. & Ours \\ \midrule 
			Straight & \textbf{100} & \textbf{100} & \textbf{100} & \underline{96} & 86 & \underline{98} & 84 & \textbf{100} & \textbf{100} & \textbf{100} \\
			One turn & \underline{95} & 88 & \textbf{96} & 92 & 16 & 90 & 78 & 94 & \underline{96} & \textbf{100} \\
			Navigation & \textbf{94} & 88 & \underline{90} & \underline{90} & 2 & 84 & 74 & \underline{86} & 76 & \textbf{92} \\
			Nav. Dynamic & \underline{89} & 80 & 82 & \textbf{90} & 2 & \underline{82} & 66 & 80 & 72 & \textbf{92}\\ \bottomrule
		\end{tabular}
	\end{adjustbox}
	
	\begin{adjustbox}{max width=.8\linewidth,center}
		\begin{tabular}{@{}ccccc||cccccc@{}}
			\toprule
			\multicolumn{11}{c}{New Town }\\ \midrule
			Task & $MP^{\segm}$ \cite{Alpher20} & $MT^{\segm \depth}$ \cite{Alpher25} & $CAL^{\temporal}$ \cite{Alpher23} & $EF^{\depth}$ \cite{Alpher22} & RL \cite{Alpher20} & IL \cite{Alpher20, Alpher03} & EF-RGB \cite{Alpher22} & CIRL \cite{Alpher24} & Ours no Att. & Ours \\ \midrule 
			Straight & 92 & \textbf{100} & 93 & \underline{96} & 74 & 97 & 82 &  \textbf{100} & 94 & \underline{99} \\
			One turn & 61 & \underline{81} & \textbf{82} & \underline{81} & 12 & 59 & 69 & \underline{71} & 37 & \textbf{79}\\
			Navigation & 24 & \underline{72} & 70 & \textbf{90} & 3 & 40 & \textbf{63} & \underline{53} & 25 & \underline{53}\\
			Nav. Dynamic & 24 & 53 & \underline{64} & \textbf{87} & 2 & 38 & \textbf{57} & \underline{41} & 18 & 40\\ \bottomrule
		\end{tabular}
	\end{adjustbox}
	
	\begin{adjustbox}{max width=.8\linewidth,center}
		\begin{tabular}{@{}ccccc||cccccc@{}}
			\toprule
			\multicolumn{11}{c}{New weather and new Town} \\ 
			\midrule
			Task & $MP^{\segm}$ \cite{Alpher20} & $MT^{\segm \depth}$ \cite{Alpher25} & $CAL^{\temporal}$ \cite{Alpher23} & $EF^{\depth}$ \cite{Alpher22} & RL \cite{Alpher20} & IL \cite{Alpher20, Alpher03} & EF-RGB \cite{Alpher22} & CIRL \cite{Alpher24} & Ours no Att. & Ours \\ \midrule 
			Straight & 50 & \textbf{96} & \underline{94} & \textbf{96} & 68 & 80 & 84 & \underline{98} & 92 & \textbf{100} \\
			One turn & 50 & \underline{82} & 72 & \textbf{84} & 20 & 48 & 76 & \underline{82} & 52 & \textbf{88} \\
			Navigation & 47 & \underline{78} & 68 & \textbf{90} & 6 & 44 & 56 & \textbf{68} & 52 & \underline{67} \\
			Nav. Dynamic & 44 & 62 & \underline{64} & \textbf{94} & 4 & 42 & 44 & \underline{62} & 36 & \underline{56} \\ \bottomrule
		\end{tabular}
	\end{adjustbox}
	\caption{Comparison with the state of the art, measured in percentage of completed tasks. Two variants of the proposed method are reported: the full method and a variant without attention. Additional sources of data used by a model are identified by superscripts: $\depth$ (depth), $\segm$ (semantic segmentation), $\temporal$ (temporal modeling). The best result per task is shown in bold and the second best is underlined, both for methods RGB-based and that use additional data.}
	\label{table:table4}
\end{table*}

\section{Results}
\label{results}
In our experiments, we train our model with the training data from \cite{Alpher03} and evaluate the performance using the driving benchmark of CARLA \cite{Alpher20}.
In order to compare our model with the state of the art, we considered several related works that use the CARLA banchmark \cite{Alpher20} as a testing protocol.
In Table \ref{table:table4} we use the acronyms MP, IL, RL to refer to the CARLA \cite{Alpher20} agents, respectively: Modular Pipeline agent, Imitation Learning agent and Reinforcement Learning agent. Note that IL was first detailed in \cite{Alpher03} and then tested on the CARLA benchmark in \cite{Alpher20}.
With the initials CAL, CIRL and MT we refer to the results from the works of Sauer et al. \cite{Alpher23}, Liang et al. \cite{Alpher24} and Li et al. \cite{Alpher25}.
Finally, EF indicates results from the work on multi-modal end-to-end autonomous driving system proposed by Xiao et al. \cite{Alpher22}.

When comparing results it must be taken into account which input modality is used and whether decisions are based on single frames or multiple frames. Our model bases its predictions solely on a single RGB frame. All baselines rely on RGB frames, but some use additional sources of data, namely depth and semantic segmentations. MP \cite{Alpher20} in addition to driving commands, predicts semantic segmentation. Similarly MT \cite{Alpher25} is built as a multi-task framework that predicts both segmentation and depth. While not using directly these sources of data as input, these models are trained with an additional source of supervision. EF \cite{Alpher22} feeds depth images along with RGB as input to the model. To show results comparable to ours, we report also the RGB variant which does not use depth information and we refer to it as EF-RGB. All models work emitting predictions one step at a time, with the exception of CAL \cite{Alpher23}, which is trained either with LSTM, GRU or temporal convolution to model temporal dynamics and take time into account.

Table \ref{table:average} reports the average success rate of episode across all tasks in the benchmark. 
Our method obtains state of the art results when compared to other methods that rely solely on RGB data. Overall, only EF \cite{Alpher22} is able to obtain a higher success rate but has to feed also depth data to the model. 
In fact, the success rate of its RGB counterpart (EF-RGB) has a 17\% drop, i.e. 8\% lower than our approach.

Table \ref{table:table4} shows the percentage of completed episodes for each task. Results are divided according to weather conditions and town. It can be seen that using attention to guide predictions allows the model to obtain good results in the evaluation benchmark, especially in the \textit{New weather} setting where our method achieves state of the art performance in 3 tasks out of 4.
We observe that methods using depth, i.e. MT \cite{Alpher25} and EF \cite{Alpher22}, perform very well in navigation tasks, hinting that adding this additional cue will likely improve the performance of our model too.

Moreover, compared to the other methods, our model has the advantage of being explainable thanks to the attention layer. In fact, instead of treating the architecture as a black box, we are able to understand what is guiding predictions by looking at attention weights over Regions of Interest.
Visual attention activations highlight which image features are useful in a goal oriented autonomous driving task, hence rejecting possible noisy features.

To underline the importance of attention, we report in Table \ref{table:average} and Table \ref{table:table4} also a variant of our model without attention. This considers the whole image instead of performing ROI-pooling on a set of boxes, but still maintains an identical multi-head structure to emit predictions. Adding the attention layer yields to an increase in performance for each task. The most significant improvements are for navigation tasks, especially in the \textit{Dynamic} setting, i.e. with the presence of other agents. Attention in fact helps the model to better take into account other vehicles, which is pivotal in automotive.

\begin{figure*}[h]
   \centering
   
\includegraphics[width=.24\textwidth]{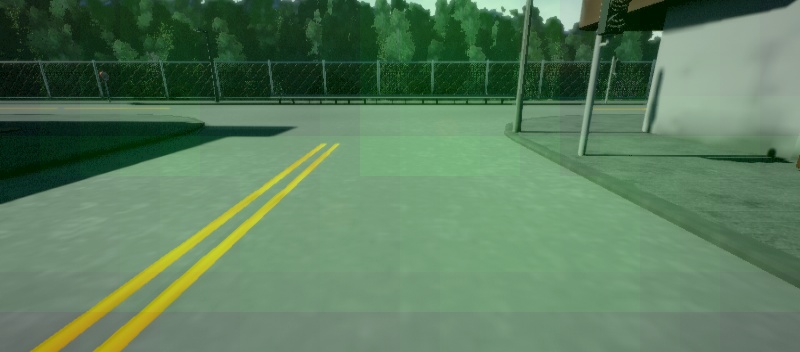}
\includegraphics[width=.24\textwidth]{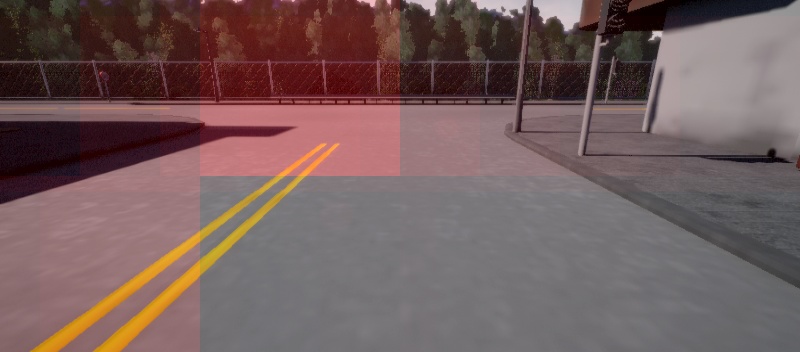}
\includegraphics[width=.24\textwidth]{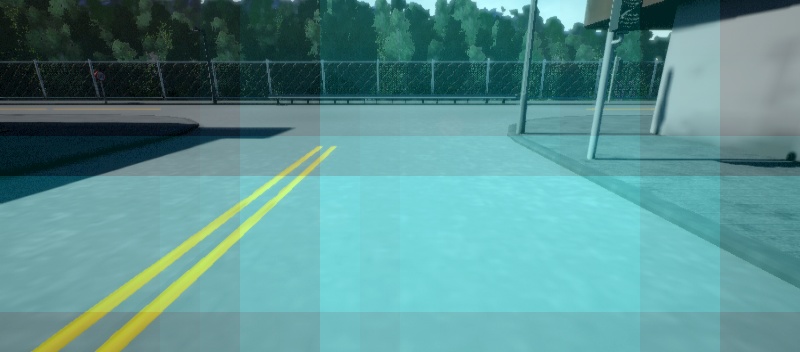}
\includegraphics[width=.24\textwidth]{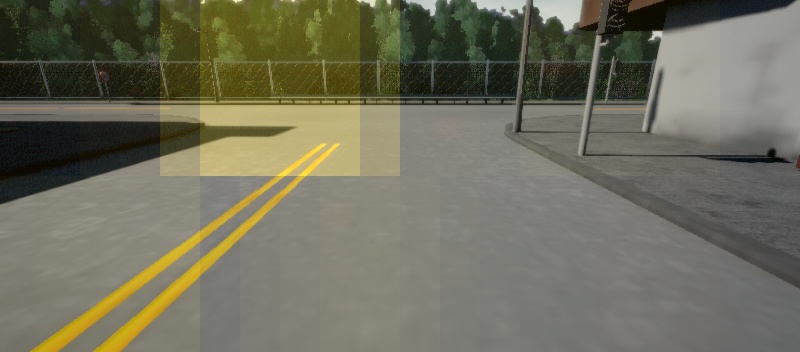}\\

\includegraphics[width=.24\textwidth]{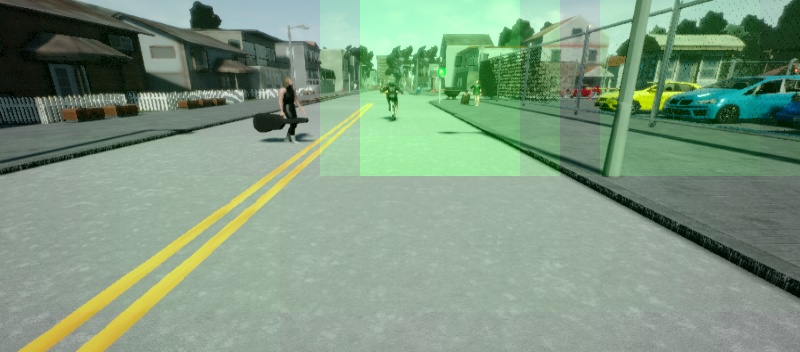}
\includegraphics[width=.24\textwidth]{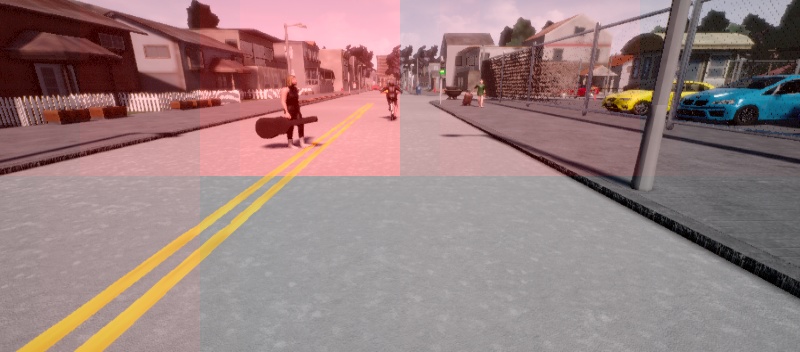}
\includegraphics[width=.24\textwidth]{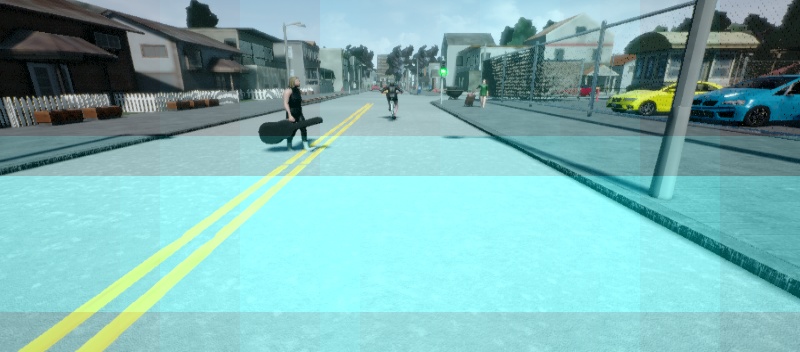}
\includegraphics[width=.24\textwidth]{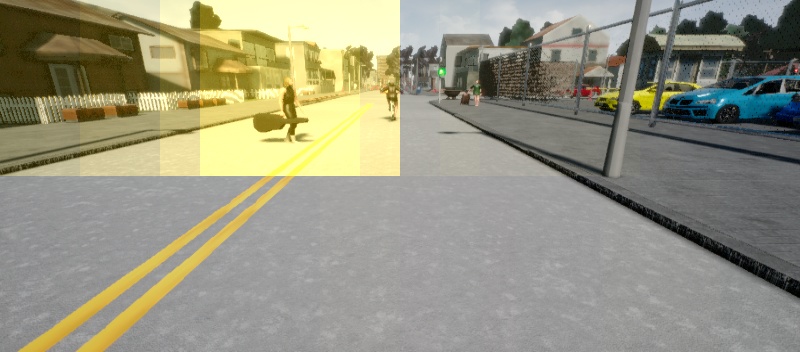}

\includegraphics[width=.24\textwidth]{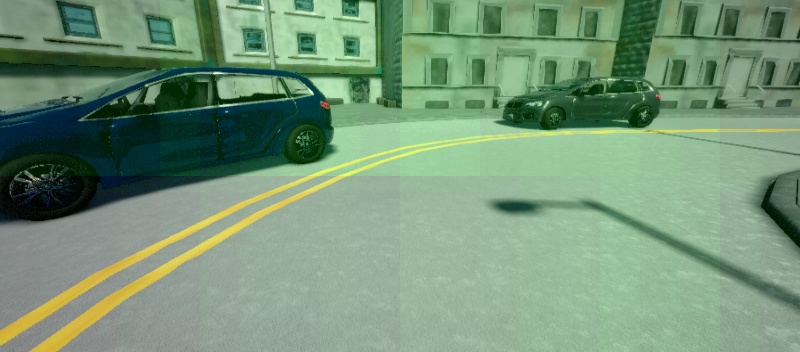}
\includegraphics[width=.24\textwidth]{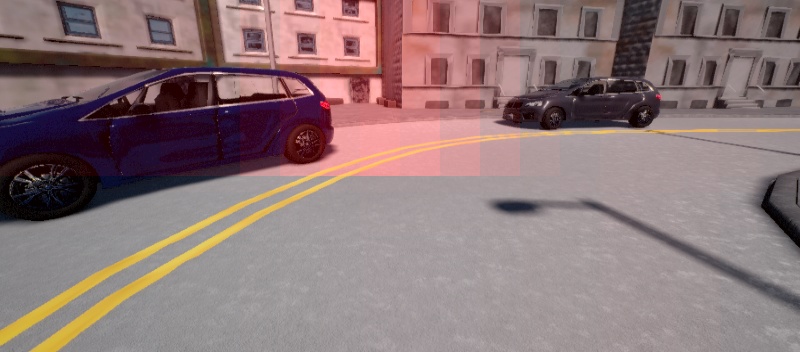}
\includegraphics[width=.24\textwidth]{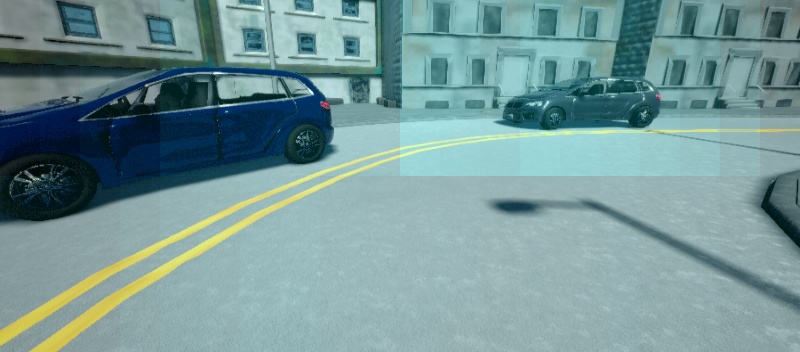}
\includegraphics[width=.24\textwidth]{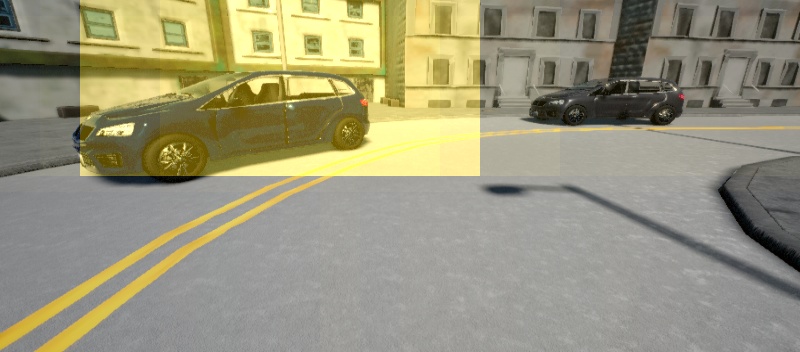}

\includegraphics[width=.24\textwidth]{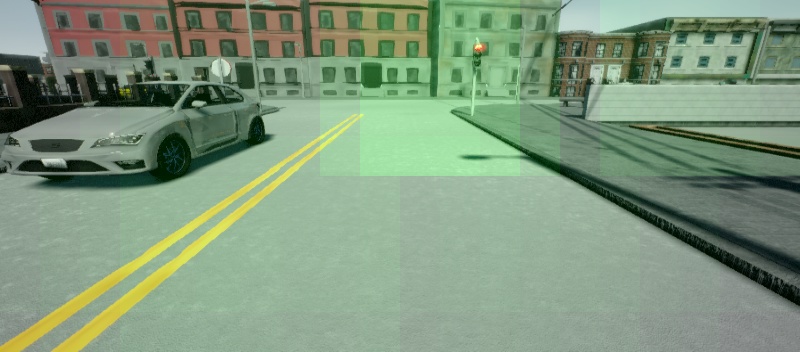}
\includegraphics[width=.24\textwidth]{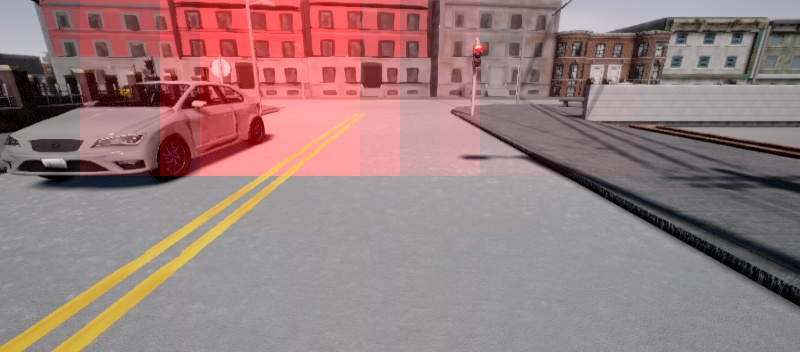}
\includegraphics[width=.24\textwidth]{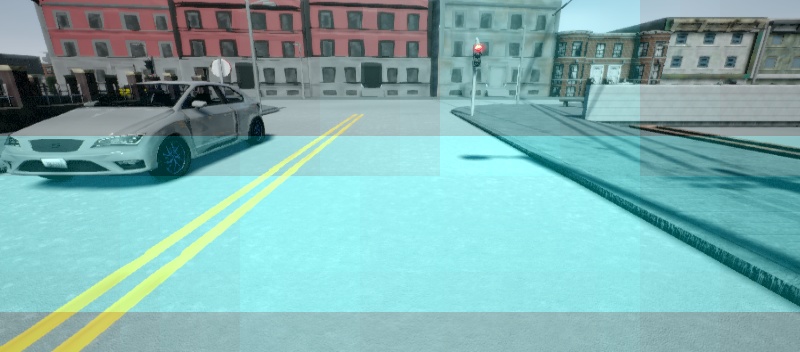}
\includegraphics[width=.24\textwidth]{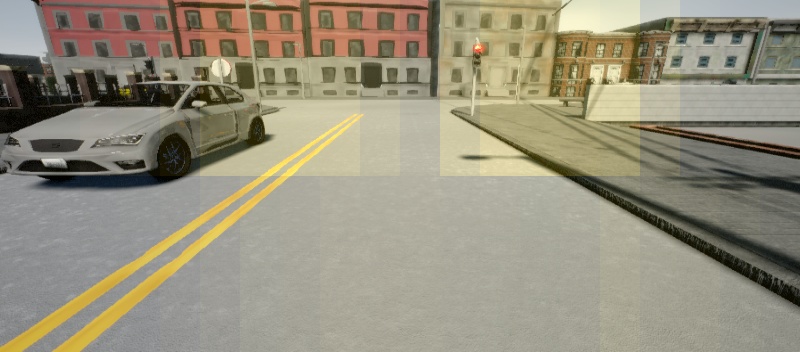}

\includegraphics[width=.24\textwidth]{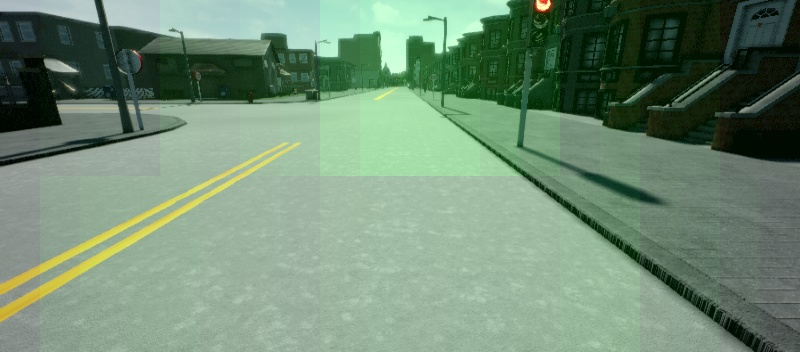}
\includegraphics[width=.24\textwidth]{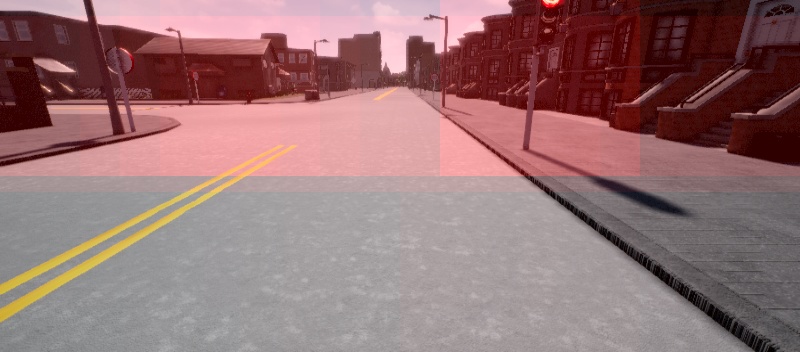}
\includegraphics[width=.24\textwidth]{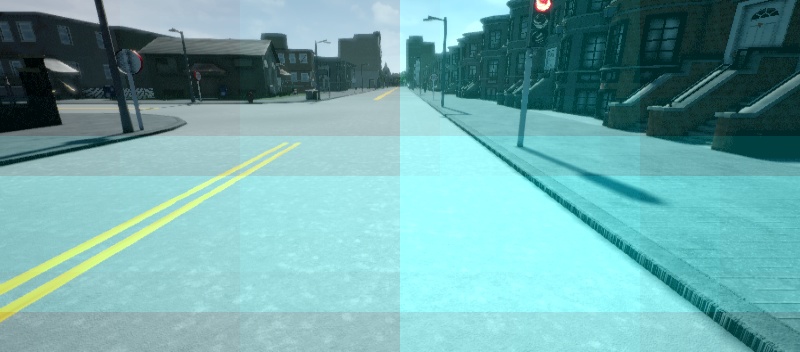}
\includegraphics[width=.24\textwidth]{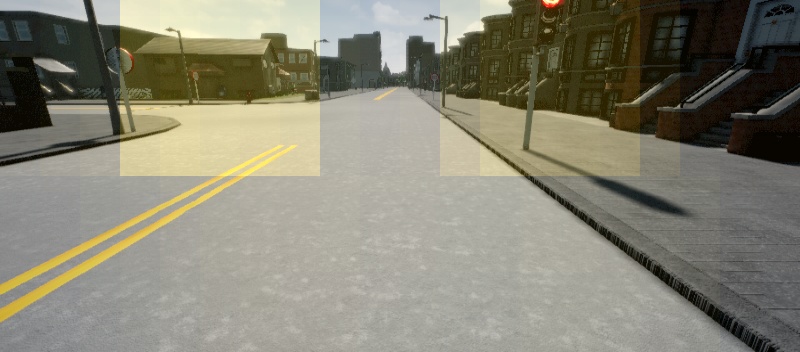}

\caption{Each row shows attention patterns on the same scene with different high level commands: \textit{Follow Lane} (green), \textit{Turn Left} (red), \textit{Turn Right} (cyan), \textit{Go Straight} (yellow).}

\label{fig:figure5}
\end{figure*}

Examples of visual attention activations for each high level command are shown in Figure \ref{fig:figure5}. It can be seen that each head of the model focuses on different parts of the scene, which are more useful for the given command. As easily imaginable, the \textit{Turn Right} and \textit{Turn Left} commands focus respectively on the right and left parts of the image. The model though, is able to identify where the intersection is, which allows the vehicle to keep moving forward until it reaches the point where it has to start turning. When a turning command is given and there is no intersection ahead, the model keeps focusing on the road, enlarging its attention area until it is able to spot a possible turning point. Interestingly, the \textit{Turn Right} usually focuses on the lower part of the image and the \textit{Turn Left} on the upper part. This is due to the fact that right turns can be performed following the right sidewalk along the curve, while for left turns the vehicle has to cross the road and merge onto the intersecting road.

The head specialized for the \textit{Go Straight} command instead focuses on both sides of the road at once, since it has to maintain an intermediate position in the lane, without turning at any intersection. The model here is focusing on the right sidewalk and the road markings to be able to keep a certain distance from both sides.

Finally, with the \textit{Follow Lane} command, the model looks ahead to understand whether the road will keep leading forward or if it will make a turn. An interesting case is given by the presence of a T-junction under the \textit{Follow Lane} command (first row in Figure \ref{fig:figure5}). This is a command that is not present in the dataset nor the evaluation benchmark, since the behavior is not well defined as the agent might turn right or left. We observed that in these cases the model stops looking only ahead and is still able to correctly make a turn in one of the two directions, without going off road.

\section{Ablation Study}
\label{sec:ablation}
\paragraph{Box type importance}
In Table \ref{table:table4} we have shown the importance of using attention in our model. We now investigate the importance of the attention boxes types.
As explained in Sec. \ref{RoiPooling}, we adopt four different box formats, varying scale and aspect ratio. We train a different model, selectively removing a targeted box type at a time. To carry out the ablation we use a subset of the CARLA benchmark composed of 10 episodes for the \textit{Straight} and \textit{One turn} tasks and 15 episodes for \textit{Navigation} and \textit{Navigation Dynamic}. Each episode is repeated for 6 different weather conditions (4 conditions can also be found in the training set and 2 are testing conditions), for a total of 300 episodes. All episodes are taken from \textit{Town1}. Results are shown in Table \ref{table:ablation}.

It emerges that all versions perform sufficiently well on simpler tasks such as \textit{Straight} and \textit{One turn}, although all models with less boxes still report a small drop in success rate. For more complex tasks though, the lack of $\textsc{big}^V$ and $\textsc{medium}$ lead to consistently worse driving performance. The box type that appears to be the most important is $\textsc{medium}$, which due to its aspect ratio and scale is able to focus on most elements in the scene. Interestingly, we observe that despite most of the boxes belong to the $\textsc{small}$ type (32 out of 48), when removing them the model is still able to obtain sufficiently high results, with a loss of 2.5 points on average. Removing these boxes though will reduce the interpretability of the model, producing much coarser attention maps.

\begin{table}[h]
\begin{adjustbox}{max width=.95\linewidth,center}
\begin{tabular}{@{}c|ccccc|c@{}}
\toprule
\midrule
Task & All & No $\textsc{big}^H$ & No $\textsc{big}^V$ & No $\textsc{medium}$ & No $\textsc{small}$ & Independent RoIs \\ \midrule 
Straight     & \textbf{100}  & \textbf{100}  & 95   & 95   & \textbf{100} & \textbf{100} \\
One turn     & \textbf{97}   & 92            & 95   & 90   & 95 & 47 \\
Navigation   & \textbf{91}   & 86            & 84   & 83   & 86 & 45 \\
Nav. Dynamic & \textbf{91}   & 88            & 81   & 83   & 88 & 39 \\ \bottomrule
\end{tabular}
\end{adjustbox}
\caption{Ablation study selectively removing a box type. Box types refer to the regions depicted in Figure \ref{fig:fig2}: $\textsc{big}^H$ (Green), $\textsc{big}^V$ (Red), $\textsc{medium}$ (Yellow), $\textsc{small}$ (Blue). We also evaluate our model with attention scores generated independently for each RoI.}
\label{table:ablation}
\end{table}

\paragraph{Fixed grid analysis}
Our model adopts a fixed grid to obtain an attention map over the image. This may look like a limitation of our architecture, restricting its flexibility. Whereas to some extent this is certainly true, designing an attention layer with a variable grid, i.e. with boxes that change in number and shape, is not trivial. Generating a variable amount of boxes, e.g. using a region proposal \cite{ren2015faster}, would require to process each box independently, depriving the attention layer of a global overview of the scene. The main issue lays in the lack of spatial information about each box: the model is indeed able to observe elements such as vehicles, traffic lights or lanes, but does not know where they belong in the image without this position being explicitly specified.

To demonstrate the inability of the model to work without a fixed grid, we modified our attention layer to emit attention scores for each RoI-pooled feature independently. This means that instead of concatenating all features and feeding them to a single dense layer, we adopt a smaller dense layer, shared across all RoIs, to predict a single attention logit. All logits are then concatenated and jointly normalized with a softmax to obtain the attention vector $\alpha$.

We show the results obtained by this model in Table \ref{table:ablation}. The only task that this architecture is able to successfully address is \textit{Straight}. As soon as the model is required to take a turn, it is unable to perform the maneuver and reach its destination. On the other hand, using a fixed grid, allows the model to learn a correlation between what is observed and where it appears in the image and jointly generating attention scores for each box. A flexible grid with variable boxes is currently an open issue and we plan to address it in future work.

\section{Acknowledgements}
\small
We thank NVIDIA for the Titan XP GPU used for this research.

\section{Conclusions}
\label{conclusion}
In this paper we have presented an autonomous driving system based on imitation learning. Our approach is equipped with a visual attention layer that weighs image regions and allows predictions to be explained. Moreover we have shown how adopting attention, the model improves its driving capabilities, obtaining state of the art results.



{\small
\bibliographystyle{ieee_fullname}
\bibliography{egbib}

\begin{thebibliography}{10}\itemsep=-1pt

\bibitem{Alpher13}
P. Anderson, X. He, C. Buehler, D. Teney, M. Johnson, S. Gould, and L. Zhang.
\newblock Bottom-up and top-down attention for image captioning and visual
  question answering.
\newblock {\em In Proceedings of the IEEE Conference on Computer Vision and
  Pattern Recognition}, pages 6077--6086, 2018.

\bibitem{Alpher05}
B.~D. Argall, S. Chernova, M. Veloso, and B. Browning.
\newblock A survey of robot learning from demonstration.
\newblock {\em Robotics and autonomous systems}, 57(5):469--483, 2009.

\bibitem{Alpher04}
A. Attia and S.Dayan.
\newblock Global overview of imitation learning.
\newblock {\em arXiv 1801.06503v1}, 2018.

\bibitem{becattini2019vehicle}
Federico Becattini, Lorenzo Seidenari, Lorenzo Berlincioni, Leonardo Galteri,
  and Alberto Del~Bimbo.
\newblock Vehicle trajectories from unlabeled data through iterative plane
  registration.
\newblock In {\em International Conference on Image Analysis and Processing},
  pages 60--70. Springer, 2019.

\bibitem{berlincioni2019road}
Lorenzo Berlincioni, Federico Becattini, Leonardo Galteri, Lorenzo Seidenari,
  and Alberto Del~Bimbo.
\newblock Road layout understanding by generative adversarial inpainting.
\newblock In {\em Inpainting and Denoising Challenges}, pages 111--128.
  Springer, 2019.

\bibitem{Alpher02}
M. Bojarski, D.~Del Testa, D. Dworakowski, B. Firner, B. Flepp, P. Goyal, L.~D.
  Jackel, M. Monfort, U. Muller, J. Zhang, X. Zhang, J. Zhao, and K. Zieba.
\newblock End to end learning for self-driving cars.
\newblock {\em arXiv preprint arXiv:1604.07316}, 2016.

\bibitem{Alpher37}
Y. Cai, D. Du, L. Zhang, L. Wen, W. Wang, Y. Wu, and S. Lyu.
\newblock Guided attention network for object detection and counting on drones.
\newblock {\em arXiv preprint arXiv:1909.113071}, 2019.

\bibitem{Alpher34}
C. Cao, X. Liu, Y. Yang, Yu Y., Z~Wang J.~Wang, Y. Huang, L. Wang, C. Huang, W.
  Xu, D. Ramanan, and T.~S. Huang.
\newblock Look and think twice: Capturing top-down visual attention with
  feedback convolutional neural networks.
\newblock {\em Proceedings of the IEEE international conference on computer
  vision}, pages 2956--2964, 2015.

\bibitem{Alpher27}
C. Chen, A. Steff, A. Kornhauser, and J. Xiao.
\newblock Deepdriving: Learning affordance for direct perception in autonomous
  drivings.
\newblock {\em In Proceedings of the IEEE International Conference on Computer
  Vision}, pages 2722--2730, 2015.

\bibitem{chen2017rethinking}
Liang-Chieh Chen, George Papandreou, Florian Schroff, and Hartwig Adam.
\newblock Rethinking atrous convolution for semantic image segmentation.
\newblock {\em arXiv preprint arXiv:1706.05587}, 2017.

\bibitem{Alpher16}
S. Chen, S. Zhang, J. Shang, B. Chen, and N. Zheng.
\newblock Brain-inspired cognitive model with attention for self-driving cars.
\newblock {\em IEEE Transactions on Cognitive and Developmental Systems}, 2017.

\bibitem{Alpher41}
Y. Chen, D. Zhao, L. Lv, and C. Li.
\newblock A visual attention based convolutional neural network for image
  classification.
\newblock {\em In 2016 12th World Congress on Intelligent Control and
  Automation (WCICA)}, pages 764--769, 2106.

\bibitem{Alpher03}
F. Codevilla, M. Müller, A. López, V. Koltun, and A. Dosovitskiy.
\newblock End-to-end driving via conditional imitation learning.
\newblock {\em In 2018 IEEE International Conference on Robotics and Automation
  (ICRA)}, pages 1--9, 2018.

\bibitem{Alpher14}
M. Cornia, L. Baraldi, and R. Cucchiara.
\newblock Show, control and tell: A framework for generating controllable and
  grounded captions.
\newblock {\em arXiv preprint arXiv:1811.10652}, 2018.

\bibitem{Alpher20}
A. Dosovitskiy, G. Ros, F. Codevilla, and A. López.
\newblock Carla: An open urban driving simulator.
\newblock {\em In Conference on Robot Learning (CoRL)}, 2017.

\bibitem{Alpher29}
H.~M. Eraqi, M.~N. Moustafa, and J. Honer.
\newblock End-to-end deep learning for steering autonomous vehicles considering
  temporal dependencies.
\newblock {\em arXiv preprint arXiv:1710.03804}, 2017.

\bibitem{geiger2013vision}
Andreas Geiger, Philip Lenz, Christoph Stiller, and Raquel Urtasun.
\newblock Vision meets robotics: The kitti dataset.
\newblock {\em The International Journal of Robotics Research},
  32(11):1231--1237, 2013.

\bibitem{Alpher30}
L. George, T. Buhet, E. Wirbel, G. Le-Gall, and X. Perrotton.
\newblock Imitation learning for end to end vehicle longitudinal control with
  forward camera.
\newblock {\em arXiv preprint arXiv:1812.05841}, 2018.

\bibitem{Alpher17}
R. Girshick.
\newblock Fast r-cnn.
\newblock {\em Proceedings of the IEEE international conference on computer
  vision}, pages 1440--1448, 2015.

\bibitem{he2017mask}
Kaiming He, Georgia Gkioxari, Piotr Doll{\'a}r, and Ross Girshick.
\newblock Mask r-cnn.
\newblock In {\em Proceedings of the IEEE international conference on computer
  vision}, pages 2961--2969, 2017.

\bibitem{Alpher11}
X.~S. Hua, L. Lu, H.~J. Zhang, and H. District.
\newblock A generic framework of user attention model and its application in
  video summarization.
\newblock {\em IEEE Transaction on multimedia}, 7(5):907--919, 2005.

\bibitem{Alpher06}
A. Hussein, M.~M. Gaber, E. Elyan, and C. Jayne.
\newblock Imitation learning: A survey of learning methods.
\newblock {\em ACM Computing Surveys (CSUR)}, 50(2):21, 2017.

\bibitem{Alpher35}
S. Jetley, N.~A. Lord, N. Lee, and P.~H. Torr.
\newblock Learn to pay attention.
\newblock {\em arXiv preprint arXiv:1804.02391}, 2018.

\bibitem{Alpher15}
K. Jinkyu and J. Canny.
\newblock Interpretable learning for self-driving cars by visualizing causal
  attention.
\newblock {\em Proceedings of the IEEE international conference on computer
  vision}, 2017.

\bibitem{Alpher19}
D.~P. Kingma and J.Ba.
\newblock Adam: A method for stochastic optimization.
\newblock {\em arXiv preprint arXiv:1412.6980}, 2014.

\bibitem{Alpher25}
Z. Li, T. Motoyoshi, T.~Ogata K.~Sasaki, and S. Sugano.
\newblock Rethinking self-driving: Multi-task knowledge for better
  generalization and accident explanation ability.
\newblock {\em in European Conference on Computer Vision (ECCV)}, 2018.

\bibitem{Alpher24}
X. Liang, T. Wang, and E.~Xing L.~Yang.
\newblock Cirl: Controllable imitative reinforcement learning for vision-based
  self-driving.
\newblock {\em in European Conference on Computer Vision (ECCV)}, 2018.

\bibitem{Alpher38}
J.~S. Lim, M. Astrid, H.~J. Yoon, and S.~I. Lee.
\newblock Small object detection using context and attention.
\newblock {\em arXiv preprint arXiv:1912.06319}, 2019.

\bibitem{Alpher40}
Y. Luo, M. Jiang, and Q. Zhao.
\newblock Visual attention in multi-label image classification.
\newblock {\em In Proceedings of the IEEE Conference on Computer Vision and
  Pattern Recognition Workshops}, 2109.

\bibitem{ma2019trafficpredict}
Yuexin Ma, Xinge Zhu, Sibo Zhang, Ruigang Yang, Wenping Wang, and Dinesh
  Manocha.
\newblock Trafficpredict: Trajectory prediction for heterogeneous
  traffic-agents.
\newblock In {\em Proceedings of the AAAI Conference on Artificial
  Intelligence}, volume~33, pages 6120--6127, 2019.

\bibitem{Alpher12}
Y.~F. Ma, L. Lu, H.J. Zhang, and M. Li.
\newblock A user attention model for video summarization.
\newblock {\em In Proceedings of the tenth ACM international conference on
  Multimedia}, pages 532--542, 2002.

\bibitem{marchetti2020memnet}
Francesco Marchetti, Federico Becattini, Lorenzo Seidenari, and Alberto
  Del~Bimbo.
\newblock Mantra: Memory augmented networks for multiple trajectory prediction.
\newblock In {\em Proceedings of the IEEE Conference on Computer Vision and
  Pattern Recognition}, 2020.

\bibitem{mur2017orb}
Raul Mur-Artal and Juan~D Tard{\'o}s.
\newblock Orb-slam2: An open-source slam system for monocular, stereo, and
  rgb-d cameras.
\newblock {\em IEEE Transactions on Robotics}, 33(5):1255--1262, 2017.

\bibitem{ren2015faster}
Shaoqing Ren, Kaiming He, Ross Girshick, and Jian Sun.
\newblock Faster r-cnn: Towards real-time object detection with region proposal
  networks.
\newblock In {\em Advances in neural information processing systems}, pages
  91--99, 2015.

\bibitem{rhinehart2019precog}
Nicholas Rhinehart, Rowan McAllister, Kris Kitani, and Sergey Levine.
\newblock Precog: Prediction conditioned on goals in visual multi-agent
  settings.
\newblock In {\em Proceedings of the IEEE International Conference on Computer
  Vision}, pages 2821--2830, 2019.

\bibitem{Alpher23}
A. Sauer, N. Savi-nov, and A. Geiger.
\newblock Conditional affordance learning for driving in urban environments.
\newblock {\em in Conference on Robot Learning (CoRL)}, 2018.

\bibitem{Alpher36}
D. Walther, U. Rutishauser, C. Koch, and P. Perona.
\newblock On the usefulness of attention for object recognition.
\newblock {\em In Workshop on Attention and Performance in Computational Vision
  at ECCV}, pages 96--103, 2004.

\bibitem{Alpher39}
F. Wang, M. Jiang, C. Qian, S. Yang., C. Li, H. Zhang, X. Wang, and X. Tang.
\newblock Residual attention network for image classification.
\newblock {\em In Proceedings of the IEEE Conference on Computer Vision and
  Pattern Recognition}, pages 3156--3164, 2107.

\bibitem{wang2017pedestrian}
Heng Wang, Bin Wang, Bingbing Liu, Xiaoli Meng, and Guanghong Yang.
\newblock Pedestrian recognition and tracking using 3d lidar for autonomous
  vehicle.
\newblock {\em Robotics and Autonomous Systems}, 88:71--78, 2017.

\bibitem{Alpher28}
Q. Wang, L. Chen, and W. Tian.
\newblock End-to-end driving simulation via angle branched network.
\newblock {\em arXiv preprint arXiv:1805.07545.}, 2018.

\bibitem{Alpher22}
Y. Xiao, F. Codevilla, A. Gurram, O. Urfalioglu, and A.~M. Lopez.
\newblock Multimodal end-to-end au-tonomous driving.
\newblock {\em arXiv preprint arXiv:1906.03199}, 2019.

\bibitem{Alpher26}
H. Xu, Y. Gao, F. Yu, and T. Darrell.
\newblock End-to-end learning of driving models from large-scale video
  datasets.
\newblock {\em In Proceedings of the IEEE conference on computer vision and
  pattern recognition}, pages 2174--2182, 2017.

\bibitem{Alpher32}
Z. Yang, J.~Yu Y.~Zhang, and J.~Luo J.~Cai.
\newblock End-to-end multi-modal multi-task vehicle control for self-driving
  cars with visual perceptions.
\newblock {\em In 2018 24th International Conference on Pattern Recognition
  (ICPR)}, pages 2722--2730, 2016.

\bibitem{Alpher09}
J. Zhang and K. Cho.
\newblock Query-efficient imitation learning for end-to-end autonomous driving.
\newblock {\em arXiv preprint arXiv:1605.06450}, 2016.

\bibitem{Alpher42}
D. Zoran, M. Chrzanowski, P.~S. Huang, S. Gowal, A. Mott, and P. Kohl.
\newblock Towards robust image classification using sequential attention
  models.
\newblock {\em arXiv preprint arXiv:1912.02184}, 2019.

\end{thebibliography}
}

\end{document}